\documentclass[runningheads]{llncs}
\usepackage[T1]{fontenc}
\usepackage{amsmath}
\usepackage{wrapfig}
\usepackage{graphicx}
\usepackage{subcaption}
\usepackage{todonotes}
\usepackage{booktabs}
\usepackage{tabularx}
\newcolumntype{Y}{>{\small\centering\arraybackslash}X}

\usetikzlibrary{decorations.pathreplacing}
\usepackage{hyperref}
\usepackage{color}

\newcommand{\refsec}[1]{Section~\ref{#1}}
\newcommand{\reffig}[1]{Figure~\ref{#1}}
\newcommand{\reftab}[1]{Table~\ref{#1}}

\newcommand{\refeq}[1]{Equation~\ref{#1}}
\newcommand{\refeqs}[1]{Eq.~\ref{#1}}

\usepackage{acronym}
\usepackage[per-mode=symbol]{siunitx}

\acrodef{IMU}{\emph{Inertial Measurement Unit}}
\acrodef{ROS}{\emph{Robot Operating System}}
\acrodef{RCLL}{\emph{RoboCup Logistics League}}
\acrodef{SLAM}{\emph{Simultaneous Localiztion And Mapping}}
\acrodef{AMCL}{\emph{Adaptive Monte Carlo Localization}}
\acrodef{MPPI}{\emph{Model Predictive Path Integral}}
\acrodef{DWB}{\emph{Dynamic Window Approach Based}}
\acrodef{LIDAR}{\emph{LIght Detection And Ranging}}
\acrodef{EKF}{\emph{Extended Kalman Filter}}
\acrodef{UKF}{\emph{Unscented Kalman Filter}}
\begin{document}
\title{A ROS~2-based Navigation and Simulation Stack for the Robotino}
%
%
\author{Saurabh Borse\inst{1} \and
Tarik Viehmann\inst{2} \and
Alexander Ferrein\inst{1} \and
Gerhard Lakemeyer\inst{2}}

\authorrunning{S. Borse et al.}
%
\institute{The Mobile Autonomous Systems and Cognitive Robotics Institute,\\ FH Aachen University of Applied Science,  52066 Aachen, Germany \and
Knowledge-Based Systems Group, RWTH Aachen University, Aachen 52074, Germany}
\maketitle              

\begin{abstract}
The Robotino, developed by Festo Didactic, serves as a versatile platform in education and research for mobile robotics tasks.
However, there currently is no ROS~2 integration for the Robotino available.
In this paper we describe our work on a Webots simulation environment for a Robotino platform extended by \ac{LIDAR} sensors. A ROS~2 integration and a pre-configured setup for localization and navigation using existing ROS packages from the Nav2 suite is provided.
We validate our setup by comparing simulations with real-world experiments conducted by three Robotinos in a logistics environment in our lab. Additionally, we tested the setup using a ROS 2 hardware driver for the Robotino developed by team GRIPS of the RoboCup Logistics League.
The results demonstrate the feasibility of using ROS~2 and Nav2 for navigation tasks on the Robotino platform showing great consistency between simulation and real-world performance.
\end{abstract}
\section{Introduction}
The Robotino is a mobile robot platform for science and education developed and distributed by Festo Didactic,
featuring a holonomic drive, close-range infrared sensors and an \ac{IMU}.
It can be used for developing navigation and mapping methods~\cite{rto_scanning19,rto_slam23} , it is being used in applications such as office mail delivery~\cite{rto_office19}, or in production logistics scenarios such as the \ac{RCLL}~\cite{RCLL2015}.

The latest Robotino platform is using Ubuntu~18.04 and 20.04 as base OS. C++ and REST APIs are provided as well as graphical programming support and \ac{ROS} 1 nodes.
However, \ac{ROS}~1~\cite{ROS} is nearing its end of life in 2025 and its successor, \ac{ROS}~2~\cite{ROS2}, offers more advanced features, including an extensive framework for navigation named as \emph{Nav2}~\cite{NAV2}.
In order to make use of this framework, several components need to be configured according to the characteristics of the robot at hand, including components for planning, path following, localization and sensing.

\begin{figure}[t]
	\centering
	\includegraphics[page=2,scale=0.75]{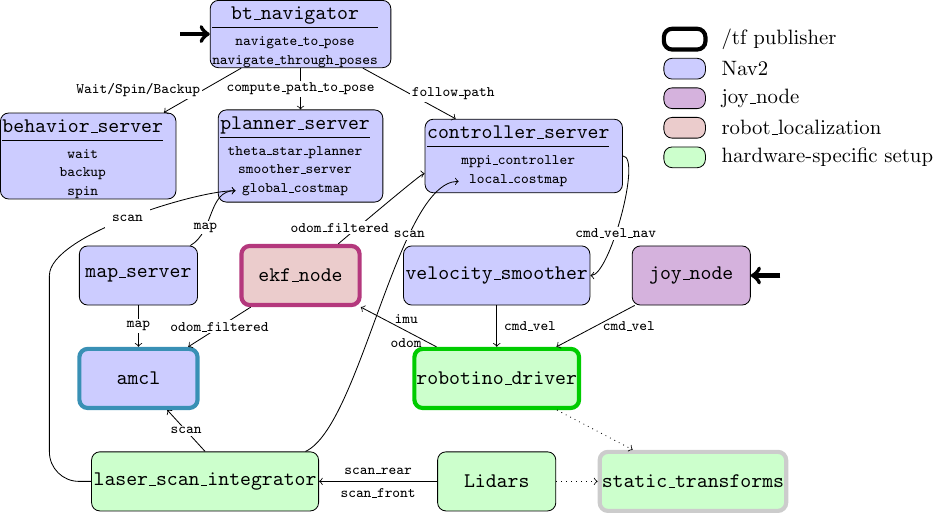}
	\caption{\ac{ROS} components overview}
	\label{fig:ros-system}
\end{figure}

In this paper, we present a \ac{ROS}~2 integration for Robotino navigation, seamlessly bridging simulation and real-world deployment for rapid prototyping and testing of navigation algorithms, by leveraging the Webots simulation framework~\cite{Webots,Webots04}.
This approach accelerates development, reduces costs, and enables reliable performance comparisons in common environments.

\begin{figure}[tbp]
    \centering
		\includegraphics[height=4cm]{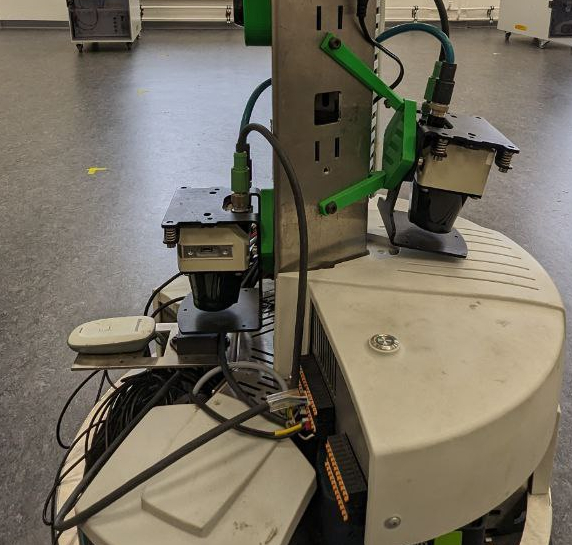}
		\includegraphics[height=4cm]{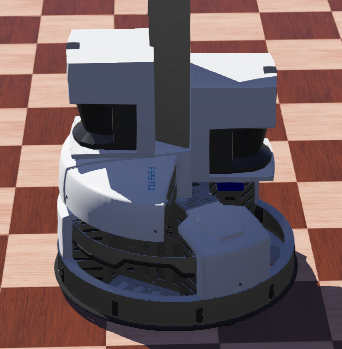}
		\caption{\ac{LIDAR} setup in simulation and on the real robot}
	\label{fig:mounts}
\end{figure}

\reffig{fig:ros-system} depicts an overview of the presented Robotino ROS~2 integration.
In order to obtain native support for the latest \ac{ROS}~2 LTS version ``Humble'', we deployed Ubuntu 22.04 and installed the core drivers from older debian packages. The installed packages are:
\emph{rec-rpc} and \emph{robotino-dev} for interprocess communication; \emph{robotino-api2} that offers a C++ interface to the hardware; and \emph{robotino-daemons} that provides the services (\emph{rpcd}, \emph{controld3} and \emph{gyrod}) to start and stop the driver.
We further extend the Robotino~4 with two SICK TiM571 \ac{LIDAR} sensors using 3D printed mounts\footnote{3D models can be found at \url{https://github.com/carologistics/hardware/tree/master/cad/robotino/stl}}as shown in \reffig{fig:mounts}.

The core component of the system is the robotino driver that models the drive kinematics and odometry of the Robotino (see \refsec{sec:kinematics}).
It translates linear and angular velocity commands, which are published over a \emph{cmd\_vel} topic given either by an input device like a joystick or by the Nav2 stack, into corresponding motor velocities.
The Robotino driver not only controls drive kinematics and odometry but also interfaces with built-in sensors like the gyroscope, infrared sensors and bumpers.
It publishes sensor data over corresponding topics and provides the static transforms for each sensor relative to the \emph{base\_link}.
Furthermore, it provides \emph{joint\_state} data vital for odometry calculations and localization.
Similarly, the external LIDARs need to provide their data in ROS, which we also account for in our simulation setup.
For the real \ac{LIDAR}s official packages provided by SICK are used to interface the data to ROS.

The rest of the paper is as follows. We present our implementation of a driver for the Webots simulation framework~\cite{Webots,Webots04} in \refsec{sec:webots-driver}.
While the Robotino platform has no official \ac{ROS}~2 driver yet, a \ac{ROS}~2 driver (using the RobotinoApi2\footnote{\url{https://wiki.openrobotino.org/index.php?title=API2}} and developed by team GRIPS from the \ac{RCLL}~\cite{GRIPS2023}) is used to interface with the hardware for our real world experiments.
%
%
The rest of the system is set up to seamlessly process data from both, simulated environments and real-world trials without a distinction of the data source.
The data from both individual \acp{LIDAR} is accumulated to a common \emph{scan} topic while an Extended Kalman Filter~\cite{Kalman1960,Kalman1970} is used to fuse the odometry and \ac{IMU} data for robust localization of the robot.
These steps provide the input for the Nav2 stack, which is used for localization and navigation as described in \refsec{sec:nav2}.
To demonstrate our results, we compare the runs of three Robotinos on a field of the \ac{RCLL} in simulation with the same runs conducted in the real world in \refsec{sec:evaluation}. Then we conclude.

%

\section{Related Work}

Pitonakova et al.~\cite{simulator_pitonakova} compare Gazebo, V-REP, and ARGoS for simulating mobile robot with differential and omnidirectional drive. They performed a number of benchmarks and evaluated the performance in terms of the Real Time Factor~(RTF) and the CPU and memory utilization. Their research concludes that Gazebo is faster for large environments, while ARGoS is able to handle more robots in small environments.

Shamshiri et al.~\cite{simulator_shamshiri} compared in their research Webots, Gazebo, ARGoS, and V-REP simulations on platforms for agricultural robotics. In their work, they compared these simulators based on performance, availability of a ROS~2 interface, a multi-physics engine, robot and sensor libraries. They come to the conclusion that V-REP fits best for their requirements and domain.


Symeonidis et al.~\cite{simulator_symeonidis} presented a comprehensive comparison between different simulators.
They highlight Gazebo and Webots for their broad community support and interoperability with \ac{ROS}~2, but see drawbacks in Gazebo regarding visual realism and computational overhead.
Meanwhile, according to their study, Webots offers fast GPU-bound rendering and a vast library support for robotic models, but has poor domain randomization tools and customization options for environments. AirSim and CARLA provide high visual realism and extensive 3D assets, but tend to be computationally intensive. Finally, CoppeliaSim offers multiple physics engines and versatile programming approaches.

As our main concern in this work is the interoperability with \ac{ROS}~2, we opted to use the Webots simulator to develop the navigation stack for the Robotino in this environment. In the following, we outline other contributions for the navigation tasks of the Robotino platform.

Zwilling et al.~\cite{zwilling2015simulation} introduce an environment created with Gazebo simulator, establishing a direct connection with the semi-autonomous game controller called referee box ensuring that it accurately mimics real-world dynamics.
However, the limitations lie in its reliance on the Fawkes framework rather than \ac{ROS}~2.

The work of Abdo et al.~\cite{robotino_sim_Abdo} focused on visual odometry and localization using the Robotino. Experiments were conducted in simulation with Robotino Sim Professional and Matlab for acquiring and processing ground truth data, as well as on actual hardware. The study evaluated visual odometry performance in challenging environments and concluded on the efficacy of visual odometry for localization.

Nizamettin et al.~\cite{robotino_nav_Nizamettin} focused on visual odometry and implemented a navigation framework using \emph{NI-LabVIEW} and the Festo navigation software stack. They evaluated the system in both simulation and real-world settings. The study identified two key limitations of visual odometry: inefficiency in extreme ambient light and difficulty detecting dynamic obstacles of similar color to the surroundings.

Bischoff et al.~\cite{robotino_nav_Bischoff} propose an hierarchical reinforcement learning (RL) architecture for mastering complex robot movements, focusing on navigation tasks with the Robotino. By decomposing movements into primitives, the approach enhances planning and execution efficiency. Results on a mobile robot platform demonstrate the efficacy of the hierarchical RL framework, with potential applications in real-time navigation and dynamic obstacle avoidance.

\section{Webots Robotino Driver}
\label{sec:webots-driver}

In order to provide a proper integration of the Robotino in Webots with \ac{ROS}~2 interfaces, several existing resources can be used, including a pre-built model of the Robotino and its sensors as well as the \ac{ROS}~2 bridge \emph{webots\_ros2} to establish seamless communication between \ac{ROS}~2 and the Webots sensor and control interfaces.

The main task is to provide a control driver that translates incoming velocity commands to motor control actions using a kinematic model of the robot as well as providing feedback about the executed actions in form of calculated odometry information and sensor readings, while exhibiting similar properties as the control interfaces of the real hardware.
In the following sections, we present the different models used in our simulation environment.

\subsection{Drive Kinematics} \label{sec:kinematics}
\begin{figure}[t]
    \centering
    \includegraphics[width=0.375\textwidth]{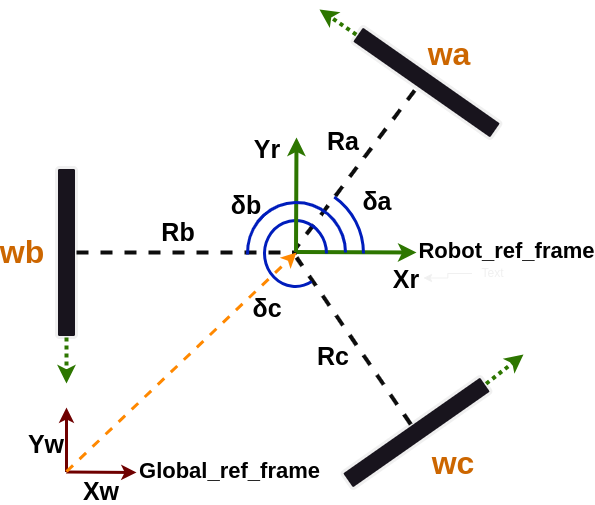}
    \caption{Representation of the pose of the robot relative to the world reference frame $(X_W, Y_W)$. The frame $(X_R, Y_R)$ is the robot frame of reference, with $X_R$ depicting its front. $(R_a, R_b, R_c)$ represents the radial distances and $(\delta_a, \delta_b, \delta_c)$ represent the angular orientation of wheels relative to the robot frame of reference}.
    \label{fig:drive_layout}
\end{figure}

The drive kinematics of the Robotino is characterized by its three side wheels enabling omnidirectional control, as depicted in \reffig{fig:drive_layout}.  The position and orientation of each wheel with respect to the robot's frame of reference play a crucial role in determining its motion and maneuverability~\cite{Kinematic_model_Palacin,Kinematic_model_Li}.

\vspace*{-\baselineskip}

\paragraph{Kinematic Model.~}
The kinematic model of the omnidirectional motion system facilitates the calculation of the rotational speeds $(\omega_{M1}, \omega_{M2}, \omega_{M3})$ of the three omnidirectional wheel motors needed (in rpm) to execute a given motion command $C = (v_{x}, v_{y}, \omega)$,
where, $v_{x}$ and $v_{y}$ are the translational velocity in $x$ and $y$ direction in \si{\meter\per\second}, respectively. $\omega$ is the angular velocity about $z$-axis (in \si{\radian\per\second}).
Palacín et al.~\cite{Kinematic_model_Palacin} describe the kinematic model that we use as defined in \refeq{eq:kinematic-matrix}.


\begin{equation}\label{eq:kinematic-matrix}
  \begin{bmatrix}
        \omega_{M1} \\
        \omega_{M2} \\
        \omega_{M3}
    \end{bmatrix} =
    \underbrace{\begin{bmatrix}
        - \sin(\delta_a) & \cos(\delta_a) & R_a \\
        - \sin(\delta_b) & \cos(\delta_b) & R_b\\
        - \sin(\delta_c) & \cos(\delta_c) & R_c
    \end{bmatrix}}_{:=K}\cdot
    \begin{bmatrix}
        v_x \\
        v_y \\
        \omega
    \end{bmatrix} \cdot \frac{1}{r} \cdot \frac{60}{2\pi} \cdot \frac{16}{1} \cdot S_{c}
    %
\end{equation}
$r$ represents the wheel radius (in [\si{\meter}]), $K$ the kinematic matrix and $\frac{16}{r}\frac{60}{2\pi}$ converts the resulting velocity to rpm. The robotino is equipped with motors GR~42x40,20W 
coupled with a planetary gearbox PLG~42S 
 with a reduction ratio of $16$. Additionally, to compensate for the absence of a motor model and gearbox in the simulation environment, we adjusted the linear and angular velocities by an empirically determined scaling factor, $S_{c}$ = 0.009375. This adjustment aimed to mimic the behavior of the real robot accurately. We determined the scaling factor by issuing identical velocity commands via the \emph{cmd\_vel} topic to both the physical robot and the simulated one in an open field, covering identical distances, and then comparing the travel times required in each case.


\paragraph{Inverse-Kinematic Model.~}
The inverse kinematic model of the omnidirectional motion system enables the determination of the motion parameters $C = (v_{x}, v_{y}, \omega)$ based on the measured actual rotational speed of the motors $(\omega_{M1}, \omega_{M2}, \omega_{M3})$.
The robotino is equipped with incremental encoders,
which facilitate measuring the rotational speed of the motors $(\omega_{M1}, \omega_{M2}, \omega_{M3})$.
These rotational speeds can be converted to the angular velocities of the wheels $(\omega_a, \omega_b, \omega_c)$. Finally, the translational velocity of the robot referred to its reference frame $v_x, v_y$ and its angular rotational velocity $\omega$ can be computed by inverting the kinematic matrix from \refeqs{eq:kinematic-matrix}, as shown in \refeqs{eq:2}.

\begin{equation}\label{eq:2}
	\begin{tikzpicture}[baseline={([yshift=0ex]current bounding box.center)}]
   \node[align=center] (a) {$
   \begin{bmatrix}
        v_x \\
        v_y \\
        \omega \\
    \end{bmatrix} = K^{-1} \cdot
    \begin{bmatrix}
        \omega_{MA} \\
        \omega_{MB} \\
        \omega_{MC}
    \end{bmatrix} \cdot r \cdot \frac{2\pi}{60} \cdot \frac{1}{16} \cdot \frac{1}{S_{c}}$
};
\end{tikzpicture}
\end{equation}



In the used Robotino model, the wheels are placed with uniform radii $(R = R_a = R_b = R_c$) and angular positions $(\delta_a = 60^\circ, \delta_b = 180^\circ, \delta_c = 300^\circ)$ as depicted in \reffig{fig:drive_layout}.

\subsection{Odometry}

The odometry we deploy for the simulated Robotino relies on the inverse kinematic analysis shown in \refsec{sec:kinematics} to estimate the robot's motion parameters $(v_x, v_y, \omega)$ relative to the its frame of reference given the motor feedback.
Following a time interval $\Delta T$, the robot's incremental position $(\Delta x, \Delta y, \Delta z)$ is determined by integrating the motion parameters $(v_x, v_y, \omega)$.
This incremental position is then added to the previously known position $(x_{i}, y_{i}, z_{i})$ of the robot with respect to the world coordinate frame to update its current position $(x_{f}, y_{f}, z_{f})$.
Here, $T_{m}$ represents the transformation matrix, transforming the pose from the robot frame of reference to the global frame of reference.
With a sufficiently small $\Delta T$ and subject to the accuracy of the encoder data, the odometry can thus be calculated according to \refeq{eq:Invkinematic_matrix}~\cite{Kinematic_model_Palacin}.

\begin{equation}\label{eq:Invkinematic_matrix}
   \begin{bmatrix}
    x_f \\
    y_f \\
    \theta_f
    \end{bmatrix} =  \begin{bmatrix}
    x_i \\
    y_i \\
    \theta_i
    \end{bmatrix} + \underbrace{\begin{bmatrix}
    \cos(\theta_i) & -\sin(\theta_i) & 0 \\
    \sin(\theta_i) & \cos(\theta_i) & 0 \\
    0 & 0 & 1
    \end{bmatrix}}_{:=T_m} \cdot \begin{bmatrix}
   v_x \\
    v_y \\
    \omega
\end{bmatrix} \cdot \Delta T
\end{equation}


While comparing the calculated odometry in simulations with the one obtained from a real Robotino, we observed notable inaccuracies occurring in the simulation, caused by unusual wheel slippage and inaccuracies due to time discretization in the simulation engine.
We therefore also implemented an alternative method to obtain the odometry in simulations, by adding a GPS sensor to the model that directly retrieves the current pose.
This alternative odometry source is used in our experiments in \refsec{sec:evaluation}.

\subsection{Sensing}
In addition to Robotino's built-in sensors such as the infrared and \ac{IMU} sensors, we extend the simulated Robotino model by SICK LMS 291 sensors, which are already defined in Webots and provide a similar coverage.
By default, Webots provides a sensor plugin that facilitates the use of sensors within the simulation environment. Through the standard \emph{webots\_ros2\_driver}, data is published over \ac{ROS} topic in every time step of the webots simulation.
Additionally, we provide static transforms of the sensors with respect to the \emph{base\_link} reference frame.


\subsection{Control}
The driver controls the movement of the Robotino in a separate act thread, which has a configurable frequency.
Whenever it receives velocity data from the \emph{cmd\_vel} topic, it stores the incoming message.
The data is used in the control loop of the act thread which converts it through the kinematics calculations of \refsec{sec:kinematics} to motor velocities and applies those velocities in simulations.
After applying the velocity, the data of the incoming message is deleted, hence the Robotino stops, if no new velocity commands arrive.
This also means that the frequency of the act loop in the driver should match the frequency of the controllers sending velocity commands.

\section{Localization and Navigation}\label{sec:nav2}
With the presented setup for acting and sensing, we utilize the \ac{ROS}~2 ecosystem to provide a basic setup for mobile robotics application by configuring a localization and navigation.

The Nav2 stack offers a node for \ac{AMCL} (see, e.g.~\cite{AMCL1999}), which is performing decently without further tuning. However, the motion model needed adaptations to fit the kinematics of the Robotino.
Additionally, as odometry information from the wheels tends to be rather inaccurate, sensor fusion is used to additionally incorporate the data from the \ac{IMU} sensor to improve the accuracy of the estimated pose.
The \emph{robot\_localization} suite in \ac{ROS} offers implementations for common sensor fusion algorithms such as the \ac{EKF} \cite{Kalman1960,Kalman1970}  and \ac{UKF}\cite{ukf_filter} algorithms.
We deploy the \emph{ekf\_node} that publishes odometry over \emph{odom\_filtered}  and also publishes the transform between \emph{odom} and \emph{base\_link}.
Meanwhile, \ac{AMCL} publishes the transform from \emph{map} to \emph{odom}. 

The \emph{nav2\_bringup} package offers sensible base configurations for planning, path following and recovery behaviors, which we took as baseline for further tuning.
We mainly focus on the planning and path following, as recovery behaviors are more likely to be tweaked to domain-specific needs.

\begin{figure}[t]
    \centering
    \begin{subfigure}[c]{0.29\textwidth}
        \includegraphics[width=\textwidth]{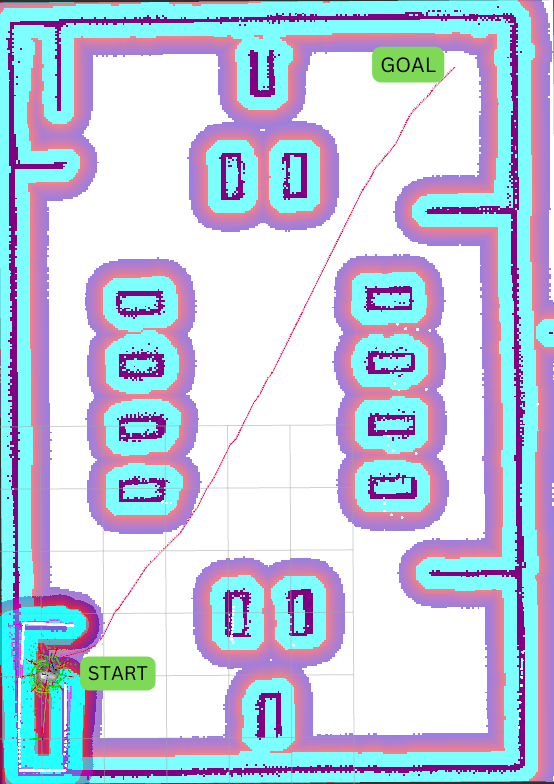}
        \caption{NavFn Planner}
        \label{fig:planners-same-goal-navfn}
    \end{subfigure}
    \begin{subfigure}[c]{0.29\textwidth}
        \includegraphics[width=\textwidth]{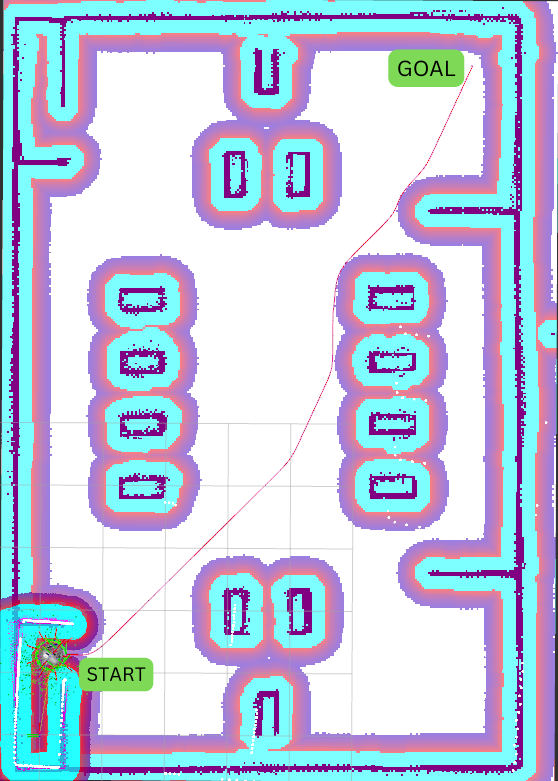}
        \caption{SmacPlanner 2D}
        \label{fig:planners-same-goal-smac}
    \end{subfigure}
    \begin{subfigure}[c]{0.29\textwidth}
        \includegraphics[width=\textwidth]{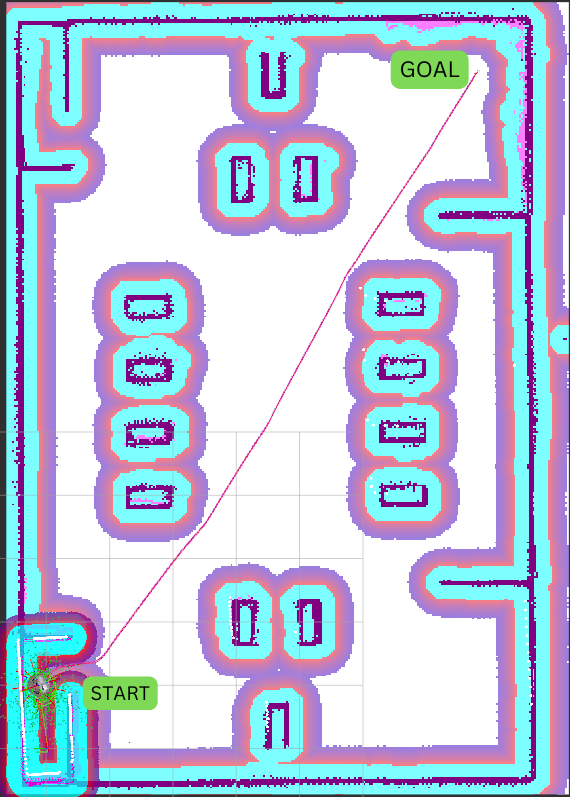}
        \caption{Theta Star Planner}
		\label{fig:planners-same-goal-theta}
    \end{subfigure}
    \caption{A comparison of the global plans generated by different planner plugins for the same goal.}
    \label{fig:planners-same-goal}
\end{figure}

The \emph{planner server} within Nav2 implements the server responsible for managing planner requests and accepts inputs such as the goal location and the name of the desired planner plugin.
%
We compare the three planner plugins that Nav2 offers for omnidirectional robots, namely \emph{NavFn}, \emph{Smac} and \emph{Theta Star}.
In \reffig{fig:planners-same-goal} shows one of the instances. The \emph{Theta Star} planner uses A* search with line of sight (LOS) checks to create any-angle paths, avoiding typical zig-zag patterns. Thus, it generates more direct trajectories compared to other planners and is the most suitable choice for this application.



%

The Controller Server in the Nav2 Stack takes care of the low-level motion control to move the robot to a desired goal pose.
The default configuration suggests to use a basic progress and goal checker as well as a \ac{DWB} \cite{NAV2} controller as a path follower.
While the \ac{DWB} controller is reactive and uses a constant action model, the \ac{MPPI} controller \cite{MPPI2016} is an alternative, which uses model predictive control to adjust trajectories on-the-fly, instead of splitting the task into a planning and execution stage.

Due to our objective of being able to navigate in environments with frequent dynamic obstacles (such as other robots), we opted for the \ac{MPPI} controller, which is more flexible compared to the \ac{DWB} controller due to it's optimization-based trajectory planning.
The \ac{MPPI} controller is particularly advantageous for its ability to predict future states of the robot using a dynamic model and optimize control actions over a finite time horizon.
This predictive capability allows the controller to anticipate obstacles and dynamically adjust trajectories, making it well-suited for navigating through complex and dynamic environments.

\begin{table}[t]
  \centering
  \caption{Main MPPI controller parameters}\label{tab:mppi-params}
  \begin{tabularx}{\linewidth}{YYYYYY} 
  \toprule
  \multicolumn{6}{l}{Parameters and values}\\
  \midrule
  \texttt{time\_steps}&\texttt{model\_dt}&\texttt{frequency}&\texttt{motion\_model}&\texttt{batch\_size}&\texttt{vx\_min}\\
  80 & 0.05 & 20& Omni &2000&-0.7\\
  \midrule
  \texttt{vx\_max}&\texttt{  wz\_max}&\texttt{vy\_max} & \texttt{vx\_std} & \texttt{vy\_std}&\texttt{wz\_std}\\
  0.7&0.8&0.7&0.4&0.4&0.4\\
\bottomrule
\end{tabularx}

\end{table}

The main considerations for configuring the \ac{MPPI} controller evolve around the kinematics constraints to create a sampling distribution and the prediction horizon which depends on controller frequency, cost map size and sampling points.
The core parameters of the \ac{MPPI} controller are listed in \reftab{tab:mppi-params}.

\section{Evaluation}\label{sec:evaluation}
\begin{figure}[b]
    \centerline{\includegraphics[width=0.9\textwidth]{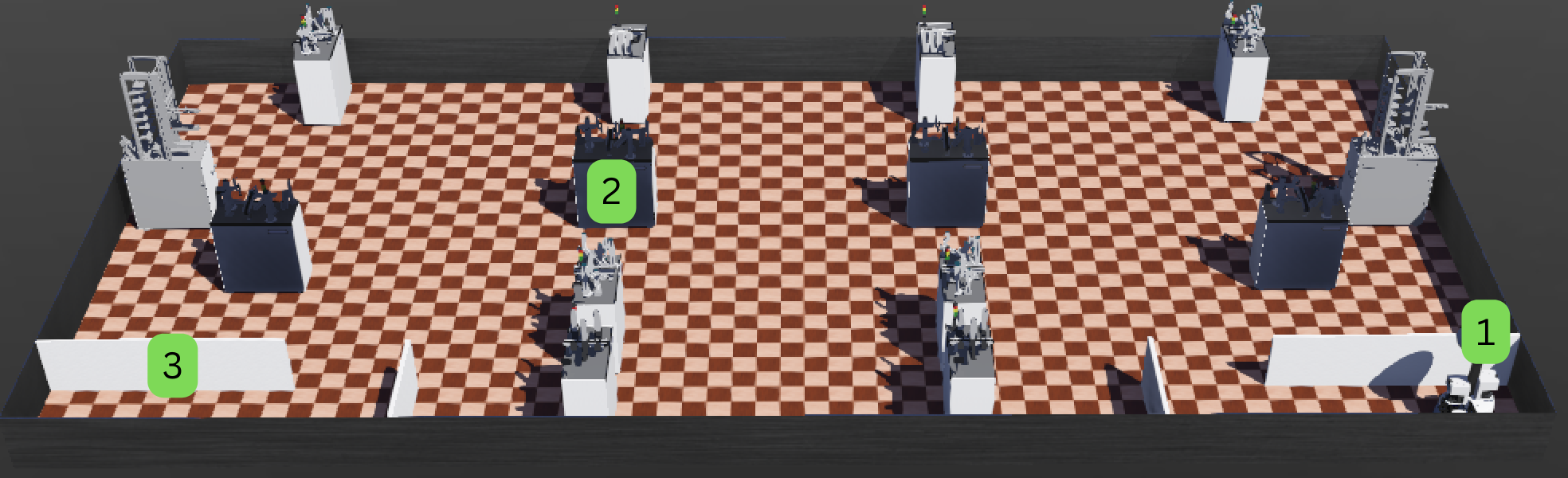}}
    \caption{Simulation map with 1 mobile robot(s), 2. static machines 3. Border walls.}
    \label{fig:map}
  \end{figure}

As a test environment, we utilize a $\SI{6}{\meter} \times \SI{12}{\meter}$ field, resembling a setup from the \ac{RCLL}, with cuboid machines serving as static obstacles (see \reffig{fig:map}).
In each experiment, robots are assigned the task of traversing five randomly generated paths. Each path consists of four waypoints, with the robots starting at the first point and then traveling to the other three points in sequential order.
The waypoints are randomly selected from points of interest in close proximity to and facing the machine sides.
Each experiment is repeated five times to gauge the consistency and \ac{ROS} 2 bags were recorded for all experiments.\footnote{\url{https://zenodo.org/records/10938688}}

In experiment (E1), a single robot was used in both simulation and the real world. In experiment (E2), all three robots were deployed for navigation trials.
\reftab{tab:exp} depicts the execution times for the paths.
We note that the ratio of total execution time in real world trials relative to simulation trials is $\approx 1.01$, hence the real world trials are about $1\%$ slower than simulation trials in this setting.

\begin{table}[tbp]
    \caption{Five distinct run per path (P) with either using one robot (E1) or using 3 robots (R1, R2 and R3) simultaneously (E2)}
    \label{tab:exp}
    \centering
    \includegraphics[page=4]{data/tables_data.pdf}
  \end{table}

\begin{figure}[htbp]
\centering
\begin{subfigure}{0.32\linewidth}
    \centering
    \includegraphics[page=1,width=\linewidth]{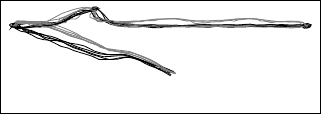}
\caption{Experiment 1 path 1}
\label{fig:e1-1}
\end{subfigure}
\begin{subfigure}{0.32\linewidth}
    \centering
    \includegraphics[page=2,width=\linewidth]{data/paths_robotino.pdf}
    \caption{Experiment 1 path 2}
\label{fig:e1-2}
\end{subfigure}
\begin{subfigure}{0.32\linewidth}
    \centering
    \includegraphics[page=8,width=\linewidth]{data/paths_robotino.pdf}
    \caption{Trial 2 path 3}
\label{fig:e2-3}
\end{subfigure}
\begin{subfigure}{0.32\linewidth}
    \centering
    \includegraphics[page=9,width=\linewidth]{data/paths_robotino.pdf}
    \caption{Trial 2 - path 4}
\label{fig:e2-4}
\end{subfigure}
\begin{subfigure}{0.32\linewidth}
    \centering
    \includegraphics[page=10,width=\linewidth]{data/paths_robotino.pdf}
    \caption{Trial 2 path 5}
    \label{fig:e2-5}
\end{subfigure}
\begin{subfigure}{0.32\linewidth}
    \centering
    \includegraphics[page=11,width=\linewidth]{data/paths_robotino.pdf}
    \caption{Trial 2 path 2, run 2}
    \label{fig:e2-2-single}
\end{subfigure}
\caption{Plots of the driven paths of each robot. Black indicates solo runs, blue, red, and green indicate paths of robots 1,2 and 3, respectively. Lighter colors are used to depict paths driven in simulation.}
\label{fig:myfig}
\end{figure}

Additionally, paths driven on the first and second experiment are plotted in \reffig{fig:e1-1} and~\ref{fig:e1-2}.
We observe that in simulations, the robot sometimes executes sharp curves, whereas trajectories in real-world trials tend to be smoother, which we attribute to the wheel slippage also observed when using inverse kinematics for pose estimation.
Also, the turning behavior in both environments could be improved (especially for real world trials), as it can involve translational movement instead of rotating on the spot or rotating while heading forward as expected.

Next, we conducted tests using three robots to gather data in dynamic environments. Each robot operates independently and perceives the others as dynamic obstacles.
The execution times are recorded in \reftab{tab:exp}.
One can observe that the real-world trials are about $16\%$ slower than simulation trials in this setting, when accumulating the execution times of all robots.
The major performance delay can be largely attributed to the increased situations, where recovery behaviors were triggered.
We conclude that this is a result of less precise position estimates (resulting from normal odometry data compared to perfect gps-based odometry, approximative static obstacle positions from being placed by humans as well as semi-transparent obstacles causing worse sensor readings).

Visualizations of the driven paths of the test instances 3, 4 and 5 are plotted in \reffig{fig:e2-3}, \ref{fig:e2-4} and \ref{fig:e2-5}.
In \reffig{fig:e2-2-single}, the  run of path 2 is depicted, which was among the worst performances due to collisions and recovery behaviors causing slow trajectories.

\section{Conclusion}
We proposed a \ac{ROS} 2 setup for a Festo Robotino extended by two \ac{LIDAR} sensors.
It provides localization and navigation on known maps using the Nav2 framework.
A Webots environment is presented that mirrors the physical setup and is used to test the framework described in this paper.
By comparing simulation trials to experiments carried out on a fleet of three Robotinos we showed
that the behavior in simulation matches the behavior in the real world, especially in environments without dynamic obstacles.
However, some unexpected wheel slippage was observed in simulation trials that is not occurring in real trials.

To build on the presented results, future work will consider the creation of custom recovery behavior, which considers costmap data to prefer driving collision-free, a possible utilization of the robots' infrared sensors for collision monitoring and leveraging bumper sensor to reduce the collision impact and aid in recovery.
Additionally, Multi Agent Path Finding (MAPF) should be explored for planning optimal collision free paths for the group of robots.
\begin{credits}
\subsubsection{\ackname}
{\footnotesize This work was partially funded by the Deutsche Forschungsgemeinschaft (DFG, German Research
Foundation) under Germany's Excellence Strategy – EXC-2023 Internet of
Production – 390621612, the EU ICT-48 2020 project TAILOR (No.\  952215) and Research Training Group 2236 (UnRAVeL)
and the Faculty of Electrical Engineering and Computer Science of FH Aachen University of Applied Sciences.
}
\end{credits}
%
%
%
\bibliographystyle{splncs04}
\bibliography{bibliography}
\end{document}